\documentclass[letterpaper, 10 pt, conference]{ieeeconf}  % Comment this line out if you need a4paper

\IEEEoverridecommandlockouts                              % This command is only needed if 
\usepackage{graphicx} % for pdf, bitmapped graphics files
\usepackage{subfigure}

\usepackage{physics}
\usepackage{amsmath} % assumes amsmath package installed
\usepackage{amssymb}  % assumes amsmath package installed
\usepackage[usenames, dvipsnames]{color}
\DeclareMathOperator{\atann}{atan2}

\title{\LARGE \bf
Quadcopter Tracking Using Euler-Angle-Free Flatness-Based Control
}

\author{Aeris El Asslouj$^{1}$ and Hossein Rastgoftar$^{2}$% <-this % stops a space
\thanks{*This work has been supported by the National Science Foundation under
Award Nos. 2133690 and 1914581.}% <-this % stops a space
\thanks{$^{1}$Aeris El Asslouj is with the  Department of Electrical and Computer Engineering,
        University of Arizona, Tucson, Arizona, USA
        {\tt\small aymaneelasslouj@arizona.edu}}%
\thanks{$^{2}$Hossein Rastgoftar is with with the Departments of  Aerospace \& Mechanical Engineering and  Electrical \& Computer Engineering,
         University of Arizona, Tucson, Arizona, USA
        {\tt\small hrastgoftar@arizona.edu}}%
}

\newcommand{\bs}[1]{
\boldsymbol{#1}
}

\newcommand{\dbs}[1]{
\dot{\boldsymbol{#1}}
}

\newcommand{\ddbs}[1]{
\ddot{\boldsymbol{#1}}
}

\newcommand{\dddbs}[1]{
\dddot{\boldsymbol{#1}}
}

\newcommand{\ddddbs}[1]{
\ddddot{\boldsymbol{#1}}
}

\begin{document}

\maketitle
\thispagestyle{empty}
\pagestyle{empty}

%%%%%%%%%%%%%%%%%%%%%%%%%%%%%%%%%%%%%%%%%%%%%%%%%%%%%%%%%%%%%%%%%%%%%%%%%%%%%%%%
\begin{abstract}
Quadcopter trajectory tracking control has been extensively investigated and implemented in the past. Available controls mostly use the Euler angle standards to describe the quadcopter's rotational kinematics and dynamics. As a result, the same rotation can be translated into different roll, pitch, and yaw angles because there are multiple Euler angle standards for characterisation of rotation in a $3$-dimensional motion space. Additionally, it is computationally expensive to convert a quadcopter's orientation to the associated roll, pitch, and yaw angles, which may make it difficult to track quick and aggressive trajectories. To address these issues,  this paper will develop a flatness-based trajectory tracking control without using Euler angles. We assess and test the proposed control's performance in the Gazebo simulation environment and contrast its functionality with the existing Mellinger controller, which has been widely adopted by the robotics and unmanned aerial system (UAS) communities.

\end{abstract}

%%%%%%%%%%%%%%%%%%%%%%%%%%%%%%%%%%%%%%%%%%%%%%%%%%%%%%%%%%%%%%%%%%%%%%%%%%%%%%%%
\section{Introduction}
Over the past few decades, multi-copter UAVs have been used for a variety of purposes, including crop management \cite{ahirwar2019application}, rescue and disaster relief missions \cite{rabta2018drone}, aerial payload transport \cite{otto2018optimization, rastgoftar2018cooperative}, surveillance \cite{mir2019drones}, piping inspections \cite{alharam2020real}. 
Trajectory tracking control of multi-copters have been extensively investigated by the researchers and multiple position-yaw controllers have been proposed. These include the cascaded Proportional–Integral–Derivative (PID) controller \cite{pid}, the Incremental Nonlinear Dynamic Inversion (INDI) controller \cite{indi}, and the Mellinger controller \cite{mellinger}.  Additionally, we recently developed a feed-back linearization-based control for quadcopter trajectory tracking \cite{Snap, rastgoftar2022real} which is called ``Snap'' controller in this paper. All these controllers are based intentionally or not on the concept of differential flatness \cite{flat} which can facilitate designing controllers for non-linear systems.  

The Mellinger controller is somewhat of an outlier in this list by the fact that it is able to follow aggressive trajectory far from the hover state while being simple and small. Meanwhile, the Snap controller offers a considerably wider domain of attraction as compared to the Mellinger controller, with the stability margin that is constrained to specific initial conditions \cite{mellinger, Snap}. While these two controllers solve the same control problem using differential flatness, they are polar opposite solutions, each with their advantages and disadvantages.

% \subsection{Contributions}
This paper compares the functionality of the  Snap and Mellinger controllers.
% \end{enumerate}
% (i) a single quadcopter tracking a predefined desired trajectory; and (ii) a quadcopter team pursuing a desired configurations in a decentralized fashion with trajectories .
% While this paper's core focus is this comparison, we offer many more contributions. These are: Reformulating the Snap controller's equations to use rotation matrices instead of Euler angles, comparing real and complex poles for tuning the controllers' gains, and establishing rotor speed limits as the main reason quadcopter controllers lose stability. The details of these contributions and their importance is presented in the rest of this section.
% This paper compares two quadcopter position-yaw controllers which are based on differential flatness. These are the Snap controller presented in \cite{Snap} and the Mellinger controller presented in \cite{mellinger}. While this paper's core focus is this comparison, we offer many more contributions. These are: Reformulating the Snap controller's equations to use rotation matrices instead of Euler angles, comparing real and complex poles for tuning the controllers' gains, and establishing rotor speed limits as the main reason quadcopter controllers lose stability. The details of these contributions and their importance is presented in the rest of this section.
Compared to the authors' previous work and existing literature, this paper offers the following main contributions:

\subsubsection{Rotation-Based Presentation of Flatness-Based Controllers}
While \cite{Snap} uses the $3-2-1$ Euler angle standard to model and control a quadcopter, this paper develops a rotation matrix-based form of the Snap controller without using Euler angles.
% We reformulated the equations of the Snap controller to use rotation matrices instead of Euler angles for defining the orientation of the quadcopter.
This is particularly beneficial because there are more than 12 Euler angles conventions not counting all variations \cite{diebel2006representing}. As a result, when implementing a controller relying on Euler angles, it is nearly always the case multiple Euler angle conventions exist for the same rotation. During our research, we found that the existing Snap controller paper \cite{Snap}, the Mellinger controller paper \cite{mellinger2011minimum}, and the Gazebo robot simulation software \cite{gazebo} each have a different convention for Euler angles requiring many conversions. However, all systems have a single convention for rotation matrices which they always provide and take as input. So the presented formulation of the Snap controller based on rotation matrices is universally compatible with all systems without a need for conversions. Note that our proposed  formulation still uses a yaw angle, but it is defined using a heading vector as opposed to an Euler angle convention.

% Meaning that when implementing any controller, it is nearly always the case that the systems involved do not have the same definition of Euler angles. During our research, we found that the Snap controller paper, the Mellinger controller paper, and the Gazebo robot simulation software we used each have a different convention for Euler angles requiring many conversions. On the other hand, all systems have a single convention for rotation matrices which they always provide and take as input. So the presented formulation of the Snap controller based on rotation matrices is universally compatible with all systems without a need for conversions. The formulation still uses a yaw angle, but it defines it using a heading vector as opposed to an Euler angle convention.

\subsubsection{Comparison of the Snap and Mellinger controllers}
We compare the Snap and Mellinger controllers both from a theoretical point of view and in terms of tracking performance during simulations. We have found that the Mellinger controller is more versatile as it can be easily converted to an attitude controller. Meanwhile, the Snap controller performs better in terms of minimizing tracking error as shown in simulations.
% We compared the Snap and Mellinger controllers both from a theoretical point of view and in terms of tracking performance during simulations. We determined that the Mellinger controller is easier to use as it requires less sensor inputs and is more versatile as it can be easily converted to an attitude controller. Meanwhile, the Snap controller performs better in terms of minimizing tracking error as shown in simulations.

\subsubsection{Comparison between complex and real poles for tuning the controllers}
We discovered that using real or complex poles does not affect the Snap Controller's performance. However, the Mellinger controller performs better when it is tuned using complex poles to a point where they are practically required. We also showed that complex poles did not lead to any non-negligible oscillations in the simulations.
%  We found that complex poles only affect the Mellinger attitude controller and greatly improve its performance to a point where they are practically required. We also showed that complex poles did not lead to any non-negligible oscillations in the simulations either with a single drone or even with a leader-follower drone network.

As a result of our study, we think that the main cause of loss of stability for quadcopter controllers is not domain of stability but rotor speed bounds. Before the domain of stability is left, there is a smaller domain out of which controllers already lose stability. This is due to the controllers attempting to get thrust and torque values that correspond to non-valid rotor speeds (negative rotor speeds or beyond maximum rotor speeds). This is an important finding because it shows that in order to make quadcopter position control more aggressive, research should focus on creating quadcopter designs which allow for negative thrust. For example, an octo-copter where 4 of its rotors create downward forces, or a quadcopter with symmetric rotors that can spin in both directions.

% \subsection{Outline}

In Section \ref{sec:problem}, we provide a problem statement and present an overview of the solution strategy for both the Snap and Mellinger controllers. In Section \ref{sec:model}, we present the dynamics of the quadcopter system using rotation matrices instead of Euler angles and define the concept of quadcopter heading. Section \ref{sec:controllers} builds on the dynamics to formulate the Snap and Mellinger controllers which are then compared from a theoretical point of view in Section \ref{sec:theoretical}. In Section \ref{sec:single_simulation}, we present the results of our simulations which are discussed in Section \ref{sec:discussion} with a conclusion in Section \ref{sec:conclusion}.

\section{Problem Statement}
\label{sec:problem}

Both the Snap and Mellinger controllers solve the quadcopter yaw-position control. They use the sensor data 
to stably track both a given smooth trajectory $\bs{r}_T$ and a given smooth yaw as a function of time $\psi_T$. More specifically, for both Mellinger and Snap controllers, the quadcopter is equipped with sensors that provide the  real-time data 
\[
\boldsymbol{s}=\left\{\boldsymbol{r},\dot{\boldsymbol{r}},\boldsymbol{R},\boldsymbol{\omega}\right\}
\]
aggregating position $\bs{r}$, velocity $\dbs{r}$, rotation matrix $\bs{R}$ specifying the orientation of the quadcopter body frame, and angular velocity $\bs{\omega}$.
% \begin{equation}
%     \boldsymbol{s} = \begin{bmatrix}\boldsymbol{r}& \dot{\boldsymbol{r}}& \boldsymbol{R}&\boldsymbol{\omega}\end{bmatrix}.
% \end{equation}
In the Quadcopter Model section, Section \ref{sec:model}, we show that rotor speeds map to thrust and torque. As such, the controllers only need to provide desired thrust and desired torque to solve the tracking problem. In other words, to ensure position-yaw tracking, Snap and Mellinger need to map $\bs{s}$, $\bs{r}_T$, and $\psi_T$ to desired thrust $p$ and desired torque $\bs{\tau}$. 
In this paper, we add the requirement that the mapping should not use Euler angles for the reasons described in the Introduction Section. 
% However, not all thrust and torque values are valid, as not all of them map to valid rotor speeds (positive and less than the maximum rotor speed). As such, if controllers provide non-valid thrust/torque values, the output rotor speeds are clamped to the valid interval. This problem is discussed in Section \ref{sec:discussion} as the main source of failure for quadcopter controllers.

As shown in the Quadcopter Control section, Section \ref{sec:controllers}, the Snap and Mellinger controllers have different strategies for solving the tracking problem. Mellinger uses a cascaded pair of position and attitude controllers. Meanwhile, Snap uses a parallel pair of position and yaw controllers. Mellinger's controllers are "cascaded" because the output of the Mellinger position controller is given as an input to the Mellinger attitude controller. Snap also has the particularity that it stores previous values of thrust and change in thrust whereas the Mellinger controller is state-less with no stored values.

The Mellinger controller's cascaded form allows it to be easily modified to become an attitude controller making it more versatile as detailed in Section \ref{sec:theoretical}. However, it seems from the results of Section \ref{sec:single_simulation} that it makes it reliant on using complex poles for tuning. The Snap controller on the other hand does not seem to benefit from complex poles and outperforms the Mellinger controller in tracking precision as shown in Section \ref{sec:single_simulation}.

\section{Quadcopter Model}
\label{sec:model}

% \subsection{Quadcopter dynamics}

We denote the inertial reference frame with base vectors $\left(\vu{i}, \vu{j}, \vu{k}\right)$ and the quadcopter's body frame with $\left(\vu{i}_B, \vu{j}_B, \vu{k}_B\right)$. $\boldsymbol{R}$ and its derivative can be expressed as:
\begin{subequations}
\begin{equation}
    \boldsymbol{R} = \begin{bmatrix}
    \vu{i}_B& \vu{j}_B& \vu{k}_B
    \end{bmatrix},
\end{equation}
\begin{equation}
    \dot{\boldsymbol{R}} = \begin{bmatrix}
    \boldsymbol{\omega} \times \vu{i}_B& \boldsymbol{\omega} \times \vu{j}_B& \boldsymbol{\omega} \times \vu{k}_B
    \end{bmatrix}.
\end{equation}
\end{subequations}

% \begin{figure}[htbp]
% \centering
% \includegraphics[width=3in]{figures/formation.png}
% \caption{Quadcopter reference pose. Red arrows show the rotation direction of each rotor.}
% \label{fig:reference_frame}
% \end{figure}

\begin{figure}
 \centering
 \subfigure[]{\includegraphics[width=0.43\linewidth]{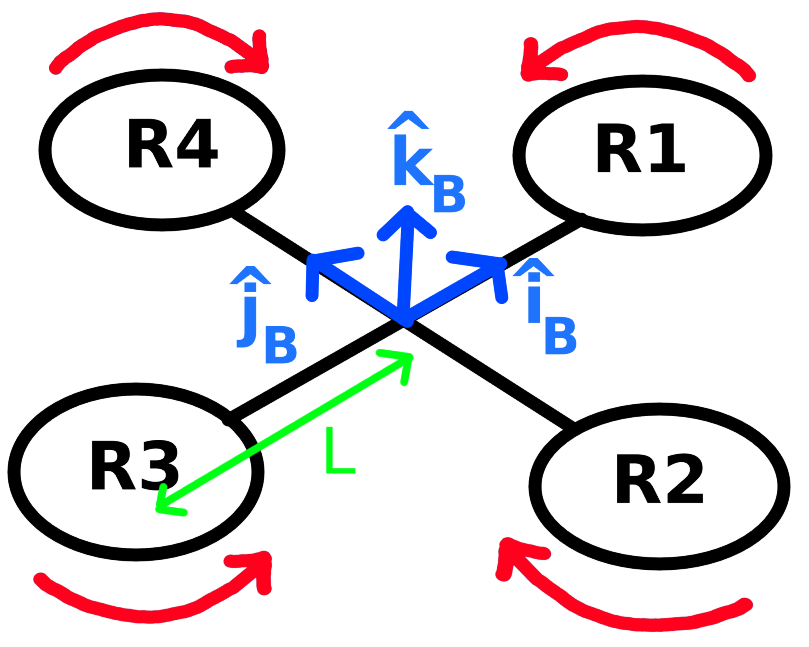}}
  \subfigure[]{\includegraphics[width=0.55\linewidth]{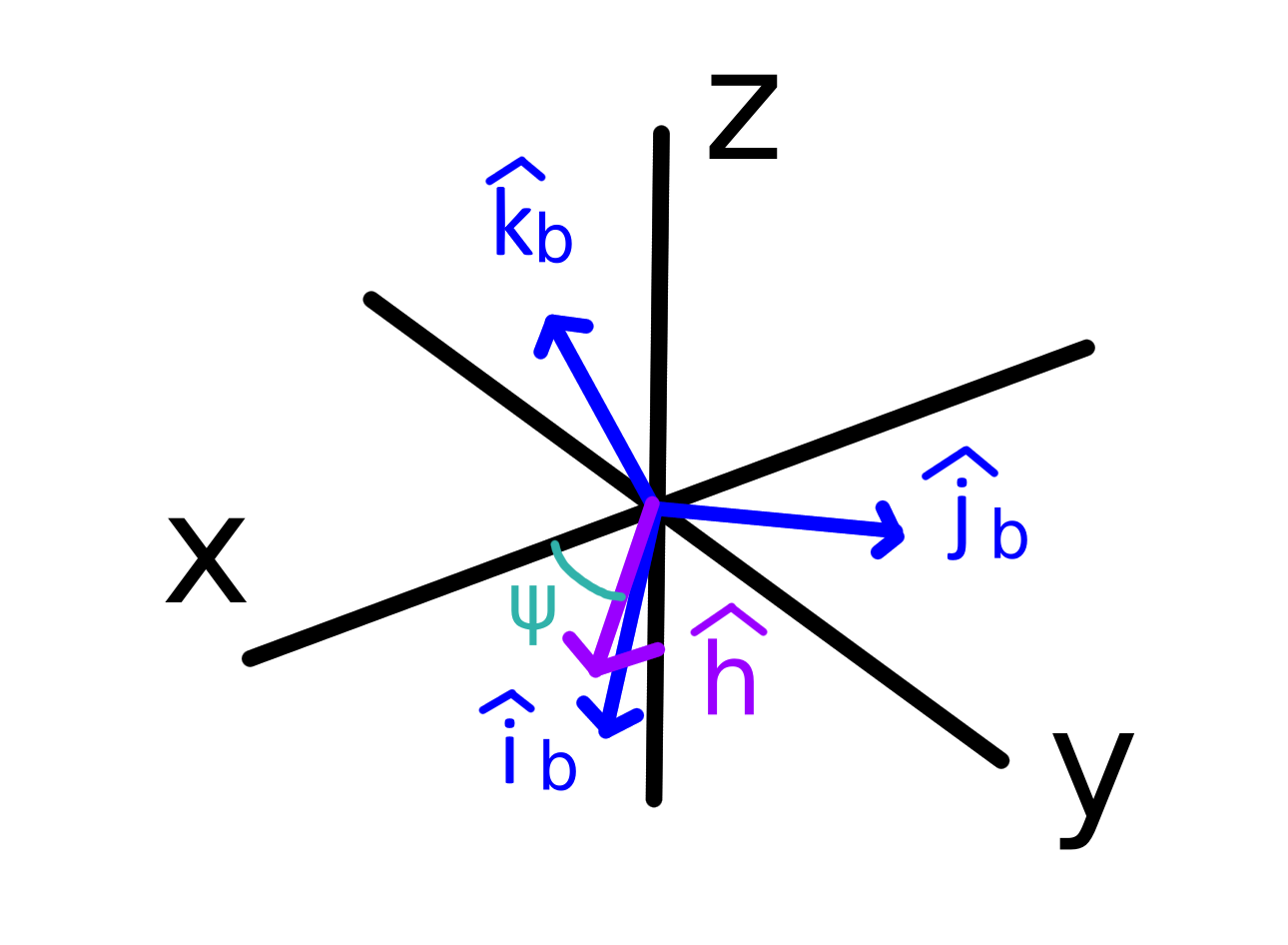}}
%  \subfigure[]{\includegraphics[width=0.32\linewidth]{Example-Initial-Formation.png}}
%  \subfigure[]{\includegraphics[width=0.32\linewidth]{Example-Final-FormationAI.png}}
%   \subfigure[$\varpi_{i,2}~\forall i\in \mathcal{V}$]{\includegraphics[width=0.9\linewidth]{SecondAngularSpeeds-V2.png}}
%   \subfigure[$\varpi_{i,3}~\forall i\in \mathcal{V}$]{\includegraphics[width=0.9\linewidth]{ThirdAngularSpeeds-V2.png}}
%   \subfigure[$\varpi_{i,3}~\forall i\in \mathcal{V}$]{\includegraphics[width=0.9\linewidth]{FourthAngularSpeeds-V2.png}}
%   \subfigure[$t=100s$]{\includegraphics[width=0.24\linewidth]{t100.jpg}}
%   \subfigure[$t=200s$]{\includegraphics[width=0.24\linewidth]{t{\color{black}400}.jpg}}
%   \subfigure[$t=400s$]{\includegraphics[width=0.24\linewidth]{t400.jpg}}
%   \subfigure[$t=500s$]{\includegraphics[width=0.24\linewidth]{t500.jpg}}
%   \subfigure[$t=650s$]{\includegraphics[width=0.24\linewidth]{t650.jpg}}
%   \subfigure[$t=812s$]{\includegraphics[width=0.24\linewidth]{t812.jpg}}
\vspace{-0.3cm}
     \caption{(a)  Quadcopter reference pose. Red arrows show the rotation direction of each rotor. (b) Heading vector and heading constraints visualized.}
\label{fig:reference_frame}
\end{figure}
The quadcopter has a mass $m$ and a diagonal inertia matrix $\boldsymbol{J}$ with entries $\left(J_x, J_y, J_z\right)$. It is setup in a ``+'' formation with a distance from rotors to center of mass $L$ as shown in Fig. \ref{fig:reference_frame} (a). Note that ``+'' formation means that the rotor arms are aligned with the body frame axes. The $i$-th rotor spins at angular speed $s_i \in \left[0, s_{max}\right]$ where $s_{max}$ is the maximum rotor angular speed which is the same for every rotor $i\in \left\{1,\cdots,4\right\}$. The $i$-th rotor creates a thrust $p_i = k_F s_i^2$ and a torque $\tau_i = \pm k_M s_i^2$ both in the $\vu{k}_B$ direction where $k_F$ and $k_M$ are aerodynamic constants. Collectively, they create a net thrust force $p \vu{k}_B$ and net torque $\boldsymbol{\tau}$ given by:
\begin{equation}
\label{eq:rotors}
    \begin{bmatrix}
    p\\
    \boldsymbol{R}^T\boldsymbol{\tau}\\
    \end{bmatrix}
    =
    \begin{bmatrix}
    k_F&k_F&k_F&k_F\\
    0&-k_FL&0&k_FL\\
    -k_FL&0&k_FL&0\\
    -k_M&k_M&-k_M&k_M
    \end{bmatrix}
    \begin{bmatrix}
    s_{1}^2\\
    s_{2}^2\\
    s_{3}^2\\
    s_{4}^2\\
    \end{bmatrix}.
\end{equation}
By applying Newton's second law, the translational and rotational dynamics of the quadcopter are given by:
\begin{subequations}
\begin{equation}
\label{eq:translation}
    m \ddot{\boldsymbol{r}} = p\vu{k}_B - mg\vu{k},
\end{equation}
\begin{equation}
\label{eq:rotational}
    \boldsymbol{J}\boldsymbol{\boldsymbol{\alpha}} + \boldsymbol{\omega}\times \left(\boldsymbol{J}\boldsymbol{\omega}\right) = \boldsymbol{\tau}.
\end{equation}
\end{subequations}
where $\boldsymbol{\alpha} = \dot{\boldsymbol{\omega}}$ is the quadcopter's angular acceleration. 
We express the body frame coordinates of $\boldsymbol{\omega}$ and $\boldsymbol{\alpha}$ with:
\begin{subequations}
\begin{equation}
    \boldsymbol{\omega} = \omega_x \vu{i}_B + \omega_y \vu{j}_B + \omega_z \vu{k}_B,
\end{equation}
\begin{equation}
    \boldsymbol{\alpha} = \alpha_x \vu{i}_B + \alpha_y \vu{j}_B + \alpha_z \vu{k}_B.
\end{equation}
\end{subequations}

\subsection{Heading model}

% \begin{figure}[htbp]
% \centering
% \includegraphics[width=3in]{figures/heading_pic.png}
% \caption{Heading vector visualized.}
% \label{fig:heading_pic}
% \end{figure}

We define the heading vector $\vu{h}$ as the normalized projection of $\vu{i}_B$ on the X-Y plane:
\begin{equation}
\label{eq:heading_def}
   \vu{h} =  \frac{(\vu{i}_B\cdot\vu{i})\vu{i} + (\vu{i}_B\cdot\vu{j})\vu{j} }{||(\vu{i}_B\cdot\vu{i})\vu{i} + (\vu{i}_B\cdot\vu{j})\vu{j}||}.
\end{equation}
Note that the denominator in Eq. \eqref{eq:heading_def} is zero only if the quadcopter is about to flip over, i.e. its front-facing vector $\vu{i}_B$ is pointing fully upward in the $\vu{k}$ direction.
Heading $\vu{h}$ is visualized in Fig. \ref{fig:reference_frame} and respects the following properties:
\begin{subequations}
\begin{equation}
\label{eq:heading_constraint_1}
    \left(\vu{k} \times \vu{h}\right).\vu{i}_B = 0,
\end{equation}
\begin{equation}
\label{eq:heading_constraint_2}
    \vu{h}.\vu{i}_B > 0.
\end{equation}
\end{subequations}
Yaw is then defined as the principal angle of $\vu{h}$ such that:
\begin{subequations}
    \begin{equation}
    \label{eq:atan}
        \psi = \atann\left(\vu{h}\cdot\vu{j}, \vu{h}\cdot\vu{i}\right),
    \end{equation}
    \begin{equation}
    \label{eq:heading}
        \vu{h} = \cos(\psi)\vu{i} + \sin(\psi)\vu{j},
    \end{equation}
\end{subequations}
where $\atann$ is the function which maps a 2d vector's y and x components to its principal angle \cite{atan2paper}. It has the property:
\begin{equation}
    \atann(A\sin(\theta), A\cos(\theta)) = \theta, \quad \forall \theta \in \left[-\pi, \pi\right], \forall A \in \mathbb{R}^{+}.
\end{equation}

\section{Quadcopter Control}
\label{sec:controllers}
Both the Snap controller and the Mellinger controller are designed to allow quadcopters to track a trajectory position $\boldsymbol{r}_T$ and trajectory yaw $\psi_T$. These approaches are presented in in Sections \ref{Snap controller} and \ref{Mellinger controller} below.

\begin{figure}[htbp]
\centering
\includegraphics[width=3.4in]{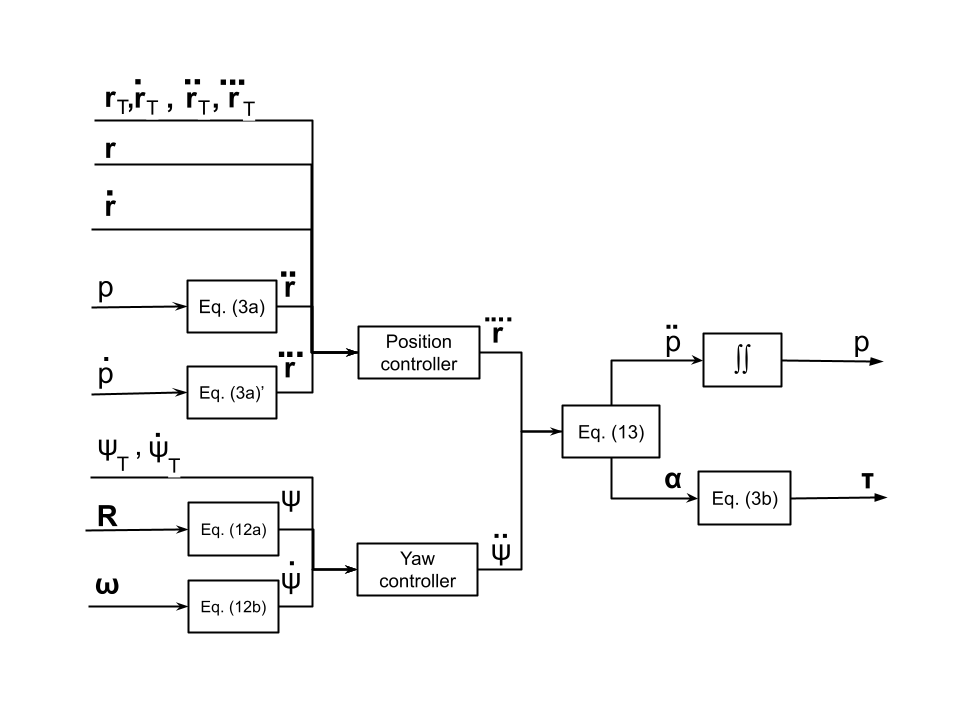}
\caption{Data flow of the Snap controller.}
\label{fig:snap_graph}
\end{figure}

\subsection{Snap controller}\label{Snap controller}

The Snap controller's data flow is visualized in Fig. \ref{fig:snap_graph}. It can be decomposed into a position controller and a yaw controller which work in parallel.

\subsubsection{Snap position controller}
To ensure position tracking, the Snap position controller tries to achieve a snap $\ddddbs{r}$ determined by:
\begin{equation}
\begin{split}
    \ddddot{\boldsymbol{r}}_{des} = \ & - \boldsymbol{K}_1 \left(\dddot{\boldsymbol{r}} - \dddot{\boldsymbol{r}}_T\right) - \boldsymbol{K}_2 \left(\ddot{\boldsymbol{r}} - \ddot{\boldsymbol{r}}_T\right) \\ - & \boldsymbol{K}_3 \left(\dot{\boldsymbol{r}} - \dot{\boldsymbol{r}}_T\right) - \boldsymbol{K}_4 \left(\boldsymbol{r} - \boldsymbol{r}_T\right).
\end{split}
\end{equation}
where $\left(\boldsymbol{K}_1, \boldsymbol{K}_2, \boldsymbol{K}_3, \boldsymbol{K}_4\right)$ are position gain matrices.

The Snap position controller needs as inputs $\bs{r}$, $\dbs{r}$, $\ddbs{r}$, and $\dddbs{r}$. $\bs{r}$ and $\dbs{r}$ are given as sensor inputs. $\ddbs{r}$ and $\dddbs{r}$ can be computed from thrust $p$ and its derivative $\dot{p}$ using the translational dynamics equation Eq. \eqref{eq:translation} and its derivative:
\begin{subequations}
\begin{equation}
    \ddot{\boldsymbol{r}} = \frac{p\vu{k}_B - mg\vu{k}}{m},
\end{equation}
\begin{equation}
    \dddot{\boldsymbol{r}} = \frac{\dot{p}\vu{k}_B - p(\boldsymbol{\omega}\times\vu{k}_B)}{m}.
\end{equation}
\end{subequations}

As $p$ and $\dot{p}$ are not sensor inputs, Snap stores their values after computing them in each control cycle to be used for the next control cycle.

% {\color{red} Please refer to Figure 2 here}

\subsubsection{Snap yaw controller}

To ensure yaw tracking, the Snap yaw controller tries to achieve a second yaw derivative $\ddot{\psi}$ determined by:
\begin{equation}
    \ddot{\psi}_{des} = - K_5 \left(\dot{\psi} - \dot{\psi}_T\right) - K_6 \left(\psi - \psi_T\right).
\end{equation}
where $\left(K_5, K_6\right)$ are positive yaw gain scalars.

The Snap yaw controller needs as inputs $\psi$ and $\dot{\psi}$. These can be obtained using:
\begin{subequations}
\begin{equation}
    \psi = \atann\left(\vu{i}_B\cdot\vu{j}, \vu{i}_B\cdot\vu{i}\right),
\end{equation}
\begin{equation}
    \dot{\psi} = \frac{\omega_k\left(\vu{k} \times \vu{h}\right) .\vu{j}_B - \omega_j \left(\vu{k} \times \vu{h}\right) .\vu{k}_B}{\vu{h} . \vu{i}_B}.
\end{equation}
\end{subequations}

\subsubsection{Mapping to thrust and torque}
Once $\ddddot{\boldsymbol{r}}_{des}$ and $\ddot{\psi}_{des}$ are obtained, Snap maps them to $\ddot{p}_{des}$ and $\boldsymbol{\alpha}_{des}$. If we derivate twice and rearrange both the translational dynamics equation Eq. \eqref{eq:translation} and the first property of the heading vector \eqref{eq:heading_constraint_1} we obtain:
\begin{subequations}
\label{eq:proj_2}
    \begin{equation}
        \vu{h}_p = m \ddddot{\boldsymbol{r}} - p \left(\boldsymbol{\omega} \times \left(\boldsymbol{\omega} \times \vu{k}_B\right)\right) - 2\dot{p}\left(\boldsymbol{\omega} \times \vu{k}_B\right),
    \end{equation}
    \begin{equation}
        \ddot{p} = \vu{h}_p.\vu{k}_B,
    \end{equation}
    \begin{equation}
        \vu{h}_\alpha = \frac{\vu{h}_p - \ddot{p}\vu{k}_B}{p},
    \end{equation}
    \begin{equation}
        \alpha_i = -\vu{h}_\alpha.\vu{j}_B,
    \end{equation}
    \begin{equation}
        \alpha_j = \vu{h}_\alpha.\vu{i}_B,
    \end{equation}
\begin{equation}
    \vu{k'} = \left(\vu{k} \times \vu{h}\right),
\end{equation}
\begin{equation}
    V = \left(\ddot{\psi}\vu{h}+\dot{\psi}^2\vu{k'}\right) . \vu{i}_B + 2\dot{\psi}\vu{h} . \left(\boldsymbol{\omega} \times \vu{i}_B\right)),
\end{equation}
\begin{equation}
\label{eq:al_k}
\alpha_k = \frac{V + \alpha_j \vu{k'} .\vu{k}_B - \vu{k'}.\left(\boldsymbol{\omega} \times \left(\boldsymbol{\omega} \times \vu{i}_B\right)\right)}{\vu{k'} .\vu{j}_B}.
\end{equation}
\end{subequations}
Note: The denominator in Eq. \eqref{eq:al_k} is zero only if the quadcopter is about to flip over. Also, these equations assume thrust $p$ is non-zero.

Then Snap integrates $\ddot{p}_{des}$ twice to get desired thrust $p_{des}$ and gets desired torque $\boldsymbol{\tau}_{des}$ from $\boldsymbol{\alpha}_{des}$ using the rotational dynamics equation Eq. \eqref{eq:rotational}.

\subsection{Mellinger controller}\label{Mellinger controller}

\begin{figure}[htbp]
\centering
\includegraphics[width=3.4in]{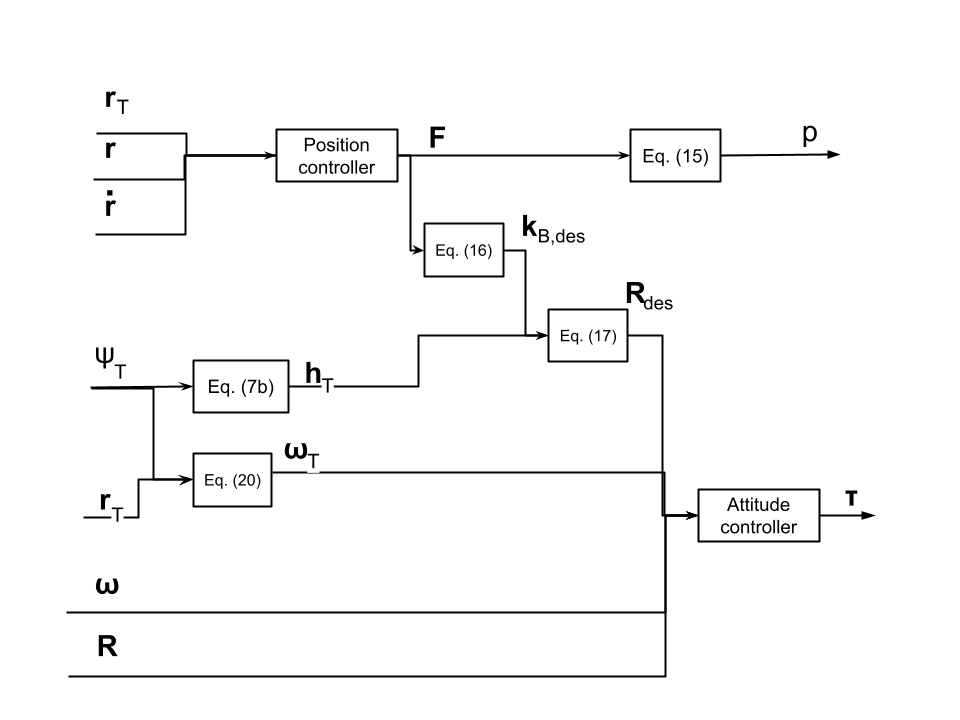}
\caption{Data flow of Mellinger controller}
\label{fig:mellinger_graph}
\end{figure}

The Mellinger controller's data flow is visualized in Fig. \ref{fig:mellinger_graph}. It is split into a position controller and an attitude controller which are cascaded with the output of the first given to the second as input.

To ensure position tracking, the Mellinger position controller tries to achieve a force determined by:
\begin{equation}
    \boldsymbol{F}_{des} = - \boldsymbol{K}_p \left(\dot{\boldsymbol{r}} - \dot{\boldsymbol{r}}_T\right) - \boldsymbol{K}_v \left(\boldsymbol{r} - \boldsymbol{r}_T\right) + mg\vu{k} + m\ddot{\boldsymbol{r}}_T,
\end{equation}
where $\boldsymbol{K}_p$ and $\boldsymbol{K}_v$ are positive definite gain matrices. The  force $\boldsymbol{F}_{des}$ is projected onto the the body frame base vector  $\vu{k}_B$ to get the desired thrust force by
\begin{equation}
    p_{des} = \boldsymbol{F}_{des}.\vu{k}_B,
\end{equation}
Additionally, we normalize $\boldsymbol{F}_{des}$ to obtain the desired orientation of $\vu{k}_B$ by
\begin{equation}\label{eq16}
    \vu{k}_{B,des} = \frac{\boldsymbol{F}_{des}}{\lVert\boldsymbol{F}_{des}\rVert}.
\end{equation}
% Note that \eqref{eq16} assume that $\lVert\boldsymbol{F}_{des}\rVert \neq 0$.

Assuming the quadcopter is not flipped over (i.e. $\vu{k}_B.\vu{k} > 0$), desired orientation vectors $\vu{i}_{B,des}$ and $\vu{j}_{B,des}$ are obtained with:
\begin{subequations}
\label{eq:base}
\begin{equation}
\label{eq:i_b}
    \vu{i}_{B,des} = \frac{\left(\vu{k} \times \vu{h}_T\right) \times \vu{k}_{B,des}}{\left|\left|\left(\vu{k} \times \vu{h}_T\right) \times \vu{k}_{B,des}\right|\right|}
\end{equation}
\begin{equation}
    \vu{j}_{B,des}  = \vu{k}_{B,des} \times \vu{i}_{B,des}
\end{equation}
\end{subequations}
Note that the denominator in Eq. \eqref{eq:i_b} is zero only if the quadcopter is about to flip over. Note that the desired body frame base vectors, denoted by $\vu{i}_{B,des}$, $\vu{j}_{B,des}$, and $\vu{k}_{B,des}$, are all obtained  form the rotation matrix $\boldsymbol{R}_{des}$ and used to compute attitude error as defined by: 
\begin{equation}
    \boldsymbol{e}_R = \frac{1}{2}\left(\boldsymbol{R}_{des}^T\boldsymbol{R}-\boldsymbol{R}^T\boldsymbol{R}_{des}\right)^\vee,
\end{equation}
where $\square^\vee$ is the vee map which maps skew-symmetric matrices to vectors:
\begin{equation}
    \begin{bmatrix}0&a&b\\-a&0&c\\-b&-c&0\end{bmatrix}^\vee = \begin{bmatrix}-c\\b\\-a\end{bmatrix}, \qquad \forall (a,b,c)\in \mathbb{R}^3.
\end{equation}
By taking the time-derivative of the translational dynamics equation Eq. \eqref{eq:translation} and the first property of the heading vector Eq. \eqref{eq:heading_constraint_1}, we obtain the components of trajectory angular velocity $\boldsymbol{\omega}_T=\omega_{i,T}\vu{k}_{B,T}+\omega_{j,T}\vu{j}_{B,T}+\omega_{k,T}\vu{k}_{B,T}$ as follows:
\begin{subequations}
\label{eq:proj_1}
\begin{equation}
    \vb{h}_\omega = \frac{m \dddot{\boldsymbol{r}}_T - \dot{p}_T\vu{k}_{B,T}}{p_T},
\end{equation}
    \begin{equation}
        \omega_{i,T} = -\vb{h}_\omega\cdot\vu{j}_{B,T},
    \end{equation}
    \begin{equation}
        \omega_{j,T} = \vb{h}_\omega.\vu{i}_{B,T},
    \end{equation}
\begin{equation}
\label{eq:om_k}
     \omega_{k,T}  = \frac{\omega_{j,T} \left(\vu{k} \times \vu{h}_T\right) .\vu{k}_{B,T} + \dot{\psi}_T\vu{h}_T . \vu{i}_{B,T}}{\left(\vu{k} \times \vu{h}_T\right) .\vu{j}_{B,T}}.
\end{equation}
\end{subequations}
We use $\boldsymbol{\omega}_T$ to compute angular velocity error as defined by:
\begin{equation}
    \boldsymbol{e}_{\omega} = \boldsymbol{\omega} - \boldsymbol{\omega}_T
\end{equation}

To ensure both yaw and position tracking, the attitude controller then tries to achieve a torque determined by:
\begin{equation}
    \boldsymbol{\tau}_{des} = -\boldsymbol{K}_R\boldsymbol{e}_R -\boldsymbol{K}_\omega \boldsymbol{e}_{\omega},
\end{equation}
where $\boldsymbol{K}_R$ and $\boldsymbol{K}_\omega$ are diagonal gain matrices.

\section{Theoretical Comparison}
\label{sec:theoretical}
In this section, we compare the stability  and versatility of the Snap and Mellinger controllers in order to assess their performance.

\subsection{Stability}
If the desired force and torques can be achieved, the domain of stability for the Snap controller is unbound as proven in \cite{Snap}. For the Mellinger controller, no proof of stability was provided, but a very similar controller was proven to be stable in \cite{geometry} given that the initial conditions respect certain constraints.

Assuming reasonable initial conditions, both controllers are able to maintain stability if they can achieve the desired thrust and torques in simulations or experiments as shown in their respective papers. What can cause loss of stability for both of them, as shown later in the simulation section, is that their desired thrust and torques cannot be achieved because they map to rotor speeds which are outside the valid range. For example, desired thrust cannot be negative or higher than what the maximum rotor speeds can provide. In these cases, the usual approach is to "clamp" or restrict the rotor speeds to the valid range by rounding them up to zero if they are negative or rounding them down to the maximum rotor speed.

Based on this, the main measures of stability performance between the two controllers is their ability to track trajectories while staying within the valid motor range and recovering from unexpected thrust/torque values caused by rotor speeds being clamped.

\subsection{Versatility}
The Mellinger controller is separated into a position controller that feeds into an attitude controller. This allows it to easily be reconfigured to an attitude controller where the position controller only controls altitude. Attitude control is useful because it allows the drone to have a manual mode where orientation and attitude, or orientation and thrust, are given by the user. On the other hand, the Snap controller cannot be modified to be used as an attitude controller without a near complete rewriting.

\begin{figure}[htbp]
\centering
\includegraphics[width=3in]{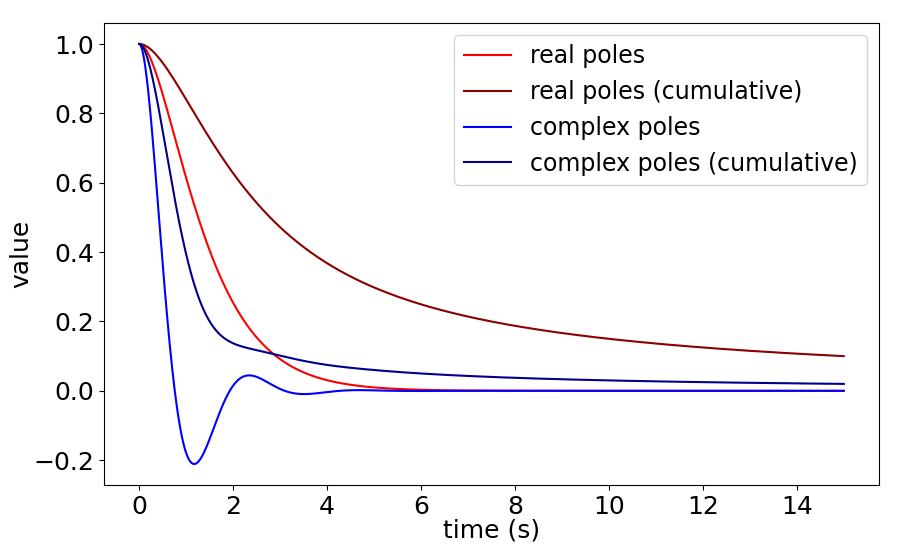}
\caption{Convergence from value 1 to value 0 with real and complex poles along with the cumulative averages}
\label{fig:pole_types}
\end{figure}

Both controllers are assumed to be tuned with pole placement \cite{poles}. For stability, the used poles need to be in the left half of the complex plane (negative real part). As shown in the simulation section, the Snap controller performs well with negative real poles, while the Mellinger controller practically requires complex poles for its attitude sub-controller. 
Complex poles allow the system to converge to the desired state faster. Convergence from value 1 to value 0 with both types of poles is shown in Fig. \ref{fig:pole_types}. As shown, the cumulative average for complex poles converges significantly faster than for real poles.

A disadvantage of complex poles is that they introduce oscillations which are undesirable for position control. However, the simulation section also shows that the oscillations from complex poles seem to be negligible.

% \subsection{Tuning}
\section{Simulation Results}
\label{sec:single_simulation}
\begin{table}[htbp]
\centering
\caption{Parameters of quadcopter model used for simulation.}
    \begin{tabular}{|c|c|c|}
    \hline
         Parameter &Value  &Unit\\
         \hline
         $m$&$0.5$  &$kg$\\
         $g$&  $9.81$ &$m/s^2$\\
         $L$&  $0.25$  &$m$\\
         $J_x$& $0.0196$ &$kg~m^2$\\
         $J_y$&$0.0196 $  &$kg~m^2$\\
         $J_z$&$0.0264$ &$kg~m^2$\\
         $k_F$& $3\times 10^{-5}$ &$N~s^2/rad^2$\\
         $k_M$&$1.1\times 10^{-6}$&$N~s^2/rad^2$\\
         \hline
    \end{tabular}
    \label{tab:quadparameters}
\end{table}
\begin{figure}[htbp]
\centering
\includegraphics[width=3in]{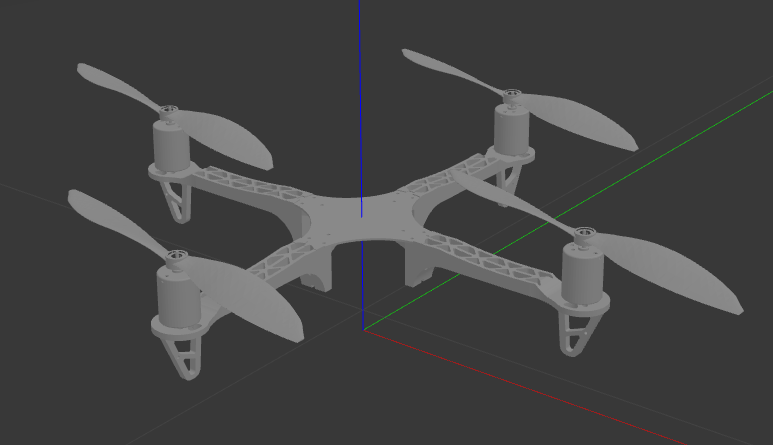}
\caption{Gazebo quadcopter model used.}
\label{fig:gazebo}
\end{figure}
To compare the Snap and Mellinger controllers. We implemented a discrete version of both in the Python scripting language with a time step of $\Delta t = 10 ms$. Then we created a simulation model in the open-source 3D robotics simulator Gazebo \cite{gazebo} based on the parameters in Table \ref{tab:quadparameters}. Fig. \ref{fig:gazebo} shows the Gazebo quadcopter model used. Finally, we bridged between the controllers and the simulation using ROS, the Robotics Operating System \cite{ros}, which allows processes to communicate.

For the tests, we used helix trajectories for a duration of $T = 10 s$ with vertical velocity of $0.1m/s$, radius $1 m$, and angular speeds $\omega \in [0 rad/s,2 rad/s]$:
\begin{equation}
    \mathbf{r}_{\omega}(t) = \begin{bmatrix}\cos(\omega \sigma_T(t))& \sin(\omega \sigma_T(t))& 0.1\sigma_T(t)\end{bmatrix},
\end{equation}
where $\sigma_T$ is an adjusted activation function defined as:
\begin{equation}
\label{eq:sigma}
    \sigma_T(t) = T\times\sigma\left(\frac{t}{T}\right) \qquad \forall t \in [0,T],
\end{equation}
where $\sigma$ is an activation function that ensures initial and final velocities, accelerations, and jerks are 0 to match the initial takeoff condition. It is defined as:
\begin{equation}
    \sigma(t) = - 20 t^7 + 70 t^6 - 84 t^5 + 35 t^4 \qquad \forall t \in [0,1]
\end{equation}

\begin{figure}[htbp]
\centering
\includegraphics[width=3in]{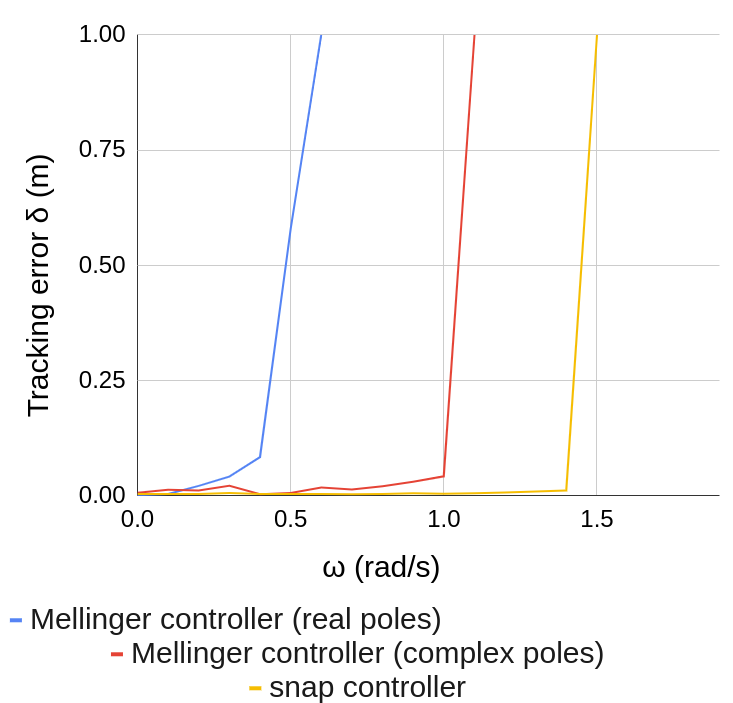}
\caption{Tracking error $\delta$ as a function of helix angular velocity $\omega$ for the Snap controller and the Mellinger controller with real and complex poles.}
\label{fig:tracking}
\end{figure}

\begin{table}[htbp]
\centering
\caption{Used poles for the Snap and Mellinger controllers.}
    \begin{tabular}{|c|c|c|}
    \hline
         Sub-controller &real poles  &complex poles\\
         \hline
         Snap: position&-10, -10, -10, -10&-\\
         Snap: yaw&  -10, -10&-\\
         Mellinger: position& -5, -5   &-5, -5\\
         Mellinger: attitude& -1, -1 &-0.5 + 3j, -0.5 - 3j\\
         \hline
    \end{tabular}
    \label{tab:poles}
\end{table}

For values of $\omega$ separated by $\Delta\omega = 0.1 rad/s$, we tested each controller in simulation and recorded the maximum tracking error $\delta$. The poles used for tuning the controllers are shown in Table \ref{tab:poles}. Fig. \ref{fig:tracking} shows the results of the tests.

\begin{figure}[htbp]
\centering
\includegraphics[width=3in]{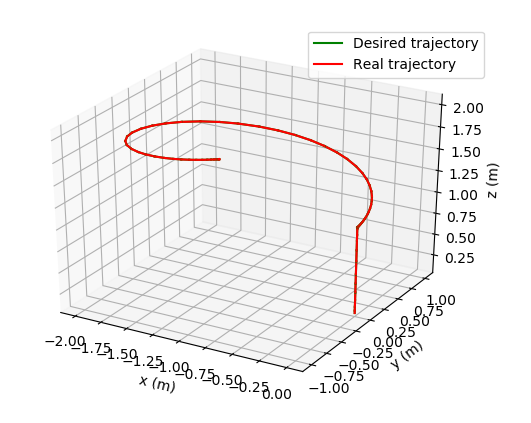}
\caption{Real (simulation) and desired trajectories for $\omega = 0.5 rad/s$ for the Snap controller. Green is desired trajectory which is hidden behind red which is real trajectory.}
\label{fig:3d_snap}
\end{figure}

\begin{figure}[htbp]
\centering
\includegraphics[width=3in]{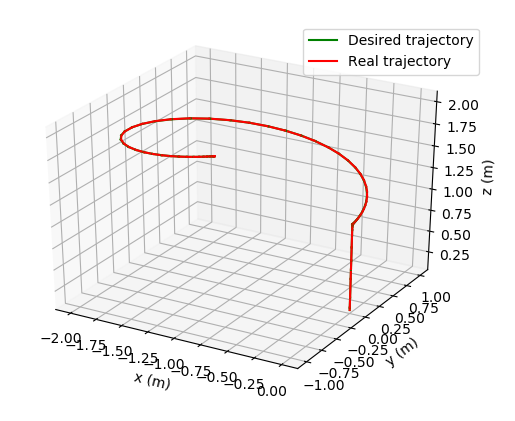}
\caption{Real (simulation) and desired trajectories for $\omega = 0.5 rad/s$ for the Mellinger controller with complex poles. Green is desired trajectory which is hidden behind red which is real trajectory.}
\label{fig:3d_mellinger}
\end{figure}

\begin{figure}[htbp]
\centering
\includegraphics[width=3in]{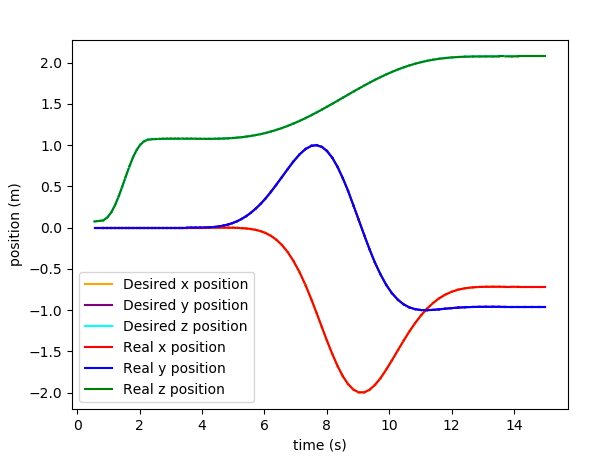}
\caption{Components of real (simulation) and desired trajectories for $\omega = 0.5 rad/s$ for the Snap controller. Desired trajectory components are hidden behind real trajectory components as error was negligible.}
\label{fig:2d_snap}
\end{figure}

\begin{figure}[htbp]
\centering
\includegraphics[width=3in]{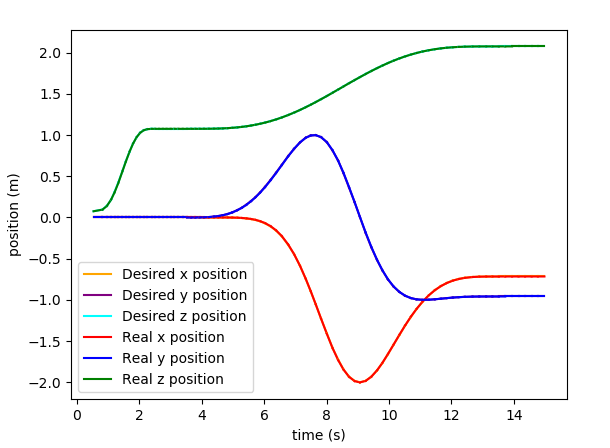}
\caption{Components of real (simulation) and desired trajectories for $\omega = 0.5 rad/s$ for the Mellinger controller with complex poles. Desired trajectory components are hidden behind real trajectory components as error was negligible.}
\label{fig:2d_mellinger}
\end{figure}

Fig. \ref{fig:3d_snap} and Fig. \ref{fig:3d_mellinger} show the real (simulation) and desired trajectories for $\omega = 0.5 rad/s$ for the Snap and Mellinger controllers (complex poles for Mellinger). Fig. \ref{fig:2d_snap} and Fig. \ref{fig:2d_mellinger} show the components of real and desired trajectories for $\omega = 0.5 rad/s$ for the Snap and Mellinger controllers (complex poles for Mellinger) as a function of time.

\section{DISCUSSION}
\label{sec:discussion}

\begin{figure}[htbp]
\centering
\includegraphics[width=3in]{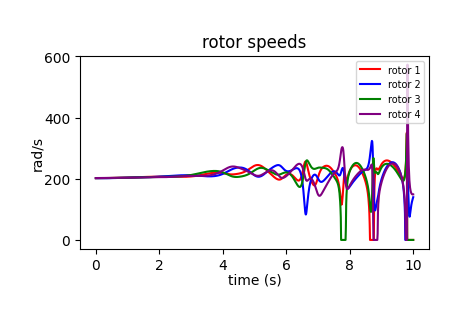}
\caption{Rotor speeds for a trajectory near the breaking point for the Mellinger controller with real poles.}
\label{fig:rotor_limit}
\end{figure}

As expected, in all cases, the tracking error became larger as $\omega$ increased. A phenomena in all cases that these graphs show is that both controllers with either sets of poles have breaking points after which they fully lose stability and spiral out of control. The reason for this seems to be controllers reaching the limits of the rotors' angular velocity range. Fig. \ref{fig:rotor_limit} shows the rotor speeds for a trajectory near the breaking point for the Mellinger controller with real poles. If the required force and torques are too high in magnitude, they require angular velocities that are too high for certain rotors (reaching the maximum rotor speed), while requiring angular velocities that are too low for the other rotors (rotor speeds cannot be negative). When this is the case, rotor speeds get clamped to the valid range leading to thrusts/torques that are different from the desired ones. This will likely make the tracking error larger which the controller will likely respond to by attempting even more out of bound forces and torques and so on. This cycle seems to lead to the loss of stability at breaking points and beyond in all cases.

For the pole types, the Mellinger controller sees a very significant increase in its performance when the poles of its attitude controller are allowed to be complex. Adding complex poles to the Mellinger position controller or to the Snap controller did not have any impact. The reason why the Mellinger controller's performance is dependent on having complex poles for its attitude controller could be because the Mellinger controller's sub-controllers are cascaded. The position controller's output is passed to the attitude controller. While the attitude controller always works, the position controller only works if the drone has the right attitude which the attitude controller is supposed to guarantee. In the case where poles are real, the attitude will exponentially decay toward the desired attitude. In the case where poles are complex, the attitude decays exponentially while oscillating around the desired attitude. The difference is the average attitude during the decay process reaches the desired faster significantly faster when oscillation is added. Using complex poles would therefore allow the position controller to function properly as the average attitude is closer to the desired attitude than with real poles.

If this is the reason complex poles affect the Mellinger attitude controller, then the same pattern should appear for all cascaded controllers with a position controller that feeds into an attitude controller. On the other hand, any non-cascaded quadcopter controllers should not be affected by complex poles.

\section{CONCLUSION}
\label{sec:conclusion}
We compared two quadcopter position-yaw controllers, the Snap controller and the Mellinger controller. We rewrote their equations to use only rotation matrices instead of Euler angles with a vector-based definition of yaw through heading. Both of them were implemented and tested on single drone simulations. We also tried both real poles and complex poles for tuning the controllers.

We found that the Snap controller outperformed the Mellinger controller in terms of minimizing tracking error. We also found that complex poles only affect the Mellinger attitude controller and greatly enhance the Mellinger controller's performance without leading to non-negligible oscillations. Our main finding is that both controllers seem to lose stability because their desired thrust and torques map to rotor speeds outside the valid range.

In terms of possible future work, there are two main problems that are still not solved. The first is the question of whether our reasoning for why complex poles affect the Mellinger attitude controller is right. This can be tested by verifying if the same behavior can be observed with other cascaded controllers. The second problem is creating a multi-copter which can create negative thrust and testing quadcopter controllers with it. This type of design would not have the rotor speed limit issue of regular quadcopter (assuming maximum rotor speeds are big enough) and should be able to perform trajectories that are significantly more aggressive than what a regular quadcopter can do.

\bibliographystyle{IEEEtran}
\bibliography{reference}

\end{document}